# HIERARCHICAL EVIDENCE AND BELIEF FUNCTIONS

Paul K. Black and Kathryn B. Laskey
Decision Science Consortium, Inc.
1895 Preston White Drive, Suite 300
Reston, Virginia 22094
(703) 620-0660

### Abstract

Dempster/Shafer (D/S) theory has been advocated as a way of representing incompleteness of evidence in a system's knowledge base. Methods now exist for propagating beliefs through chains of inference. This paper discusses how rules with attached beliefs, a common representation for knowledge in automated reasoning systems, can be transformed into the joint belief functions required by propagation algorithms. A rule is taken as defining a conditional belief function on the consequent given the antecedents. It is demonstrated by example that different joint belief functions may be consistent with a given set of rules. Moreover, different representations of the same rules may yield different beliefs on the consequent hypotheses.

## 1. Introduction

A popular way of representing knowledge in automated reasoning systems is by a set of rules, each asserting belief in some consequent hypothesis conditional on belief in some set of antecedent hypotheses. Rule-based systems are popular because they are modular (in the sense that each rule represents a single, separable "bit" of knowledge), and because the rule format appears to be a natural way for humans to encode knowledge. The early idea of rule-based systems was to process knowledge entirely symbolically, but applications soon demanded the expression of degrees of belief or certainty associated with rules.

The developers of MYCIN (Buchanan and Shortliffe, 1984) assigned "certainty factors" to rules, which propagated in an *ad hoc* manner. Adams (1984) showed the equivalence of the MYCIN model with a probabilistic model under certain (not too plausible) independence assumptions. Since that time, researchers have investigated propagation mechanisms for a number of uncertainty representations. We focus in this paper on belief functions, which have been advocated as a way of representing incompleteness of evidence (i.e., the evidence may bear on the truth of a hypothesis, but be insufficient to prove it).

A fully general model of beliefs (or probabilities) in a network of hypotheses requires a specification of a joint belief function (or probability distribution) over the entire space of possible combinations of values of all the variables. Clearly, such a model would be prohibitively difficult to assess in the absence of simplifying assumptions. The most common simplifying assumption is to assume that the directed graph representing the inferential model satisfies a Markov property (Lauritzen and Spiegelhalter, 1987; Shafer, Shenoy, and Mellouli, 1986). That is, a node is "screened" by its direct neighbors from all other nodes in the graph. The Markov property means that beliefs in a node depend on other nodes in the graph only through their effects on the beliefs in its neighbors. This assumption not only simplifies

22

Suppose in our example that an author of this paper assessed the following conditional beliefs for E given A and $\bar{A}$. (For convenience, we use A to denote A = 1, and $\bar{A}$ to denote A = 0; similarly for E.) "I feel that A justifies a degree of belief of *at least* .8 in E, but it may be higher. I don't want to assign any belief directly to $\bar{E}$." This could be expressed as a conditional belief function over E given A, that assigned belief .8 directly to E, and left the rest of the belief *uncommitted* (i.e., the remaining belief of .2 does not discriminate between E and $\bar{E}$). We would say that there is a .8 chance that A *proves* E, and a .2 chance that A has no bearing on the truth of E. The difference between a proposition's *belief* and its *plausibility* (1 minus the belief of its complement) represents the range of permissible belief. This range is .8 to 1.0 for E given A, and 0 to .2 for $\bar{E}$ given A.

Suppose further that conditional on $\bar{A}$ the author assigned belief .5 directly to E, leaving the remaining belief of .5 uncommitted with respect to E. Assume also that the incoming beliefs (whether directly assessed or propagated) assign belief .3 directly to A and .2 directly to $\bar{A}$, leaving the remaining belief of .5 uncommitted.

Having assessed the above beliefs on A, what should be our resultant beliefs on E? Shafer, Shenoy and Mellouli (1986) discuss a method analogous to Pearl's for propagating D/S beliefs in a network satisfying a Markov condition. Laskey and Lehner (1988, in press) discuss how to use assumption-based truth maintenance to propagate beliefs when the Markov condition holds. These two propagation mechanisms are formally equivalent (except that the former cannot handle nonindependencies due to shared antecedents). Both require specification of joint beliefs. Laskey and Lehner allow direct specification of beliefs as rules with "belief tokens" as antecedents, but the joint beliefs are implicitly defined in the rule specification.

To propagate beliefs in a Markov network, we represent the link A ⟶ E by a joint belief function over the cross-product space A×E. The incoming beliefs on A implicitly define a belief function over A×E that is vacuous on E (i.e., it contains information about A, but no information about which of E or $\bar{E}$ is the case). Applying Dempster's Rule produces a combined belief function over A×E, which is then collapsed into a marginal belief function over E. This resultant belief function may then serve as the input to the link E ⟶ W, just as the original beliefs in A might have been propagated through the link D ⟶ A from beliefs in D.

The problem here is how to obtain the joint belief function over the cross product space. In the case of a probabilistic model, the two conditional probability distributions given A and $\bar{A}$ are sufficient to represent the link A ⟶ E, and could be elicited by simply attaching numbers to rules. Unfortunately, there is no *unique* way of transforming the conditional belief functions we elicited into a joint belief function so that Shafer, Shenoy and Mellouli's (1986) propagation mechanism can be applied. And different ways of creating the joint beliefs result in different marginal belief functions on E. In other words, *specifying an inference "rule" (conditional belief function) from each value of A to the set E is not sufficient to uniquely determine how beliefs in A propagate to beliefs in E.*

23

propagation algorithms, but also enables the assessment process to be limited to a few hypotheses at a time.

Pearl (1986) discusses a way to add probabilistic information to a rule-based system. The simplicity and understandability of the rule-based formalism are retained in this representation. Each rule is associated with a strength, or degree to which the antecedents justify the conclusion. In a probabilistic system, this strength is represented by a conditional probability. The full joint probability distribution is completely determined by: (i) the conditional probability distributions over each consequent hypothesis set given each possible combination of values of its antecedent hypothesis sets; and (ii) the marginal probability distributions of all hypothesis sets corresponding to root nodes.

Likewise, it seems natural to assess a belief function model by assessing conditional belief functions on consequent hypothesis sets and marginal belief functions on root-node hypothesis sets. Unfortunately, unlike in the probabilistic case, this procedure is insufficient to uniquely determine a joint belief function over the entire hypothesis space (Dempster, 1968; Shafer, 1982). A full joint belief function could be specified by direct assessment (cf., Shafer, 1982), but this procedure gives up the simplicity and transparency of the rule-based representation.

If the belief function representation is to prove useful as a method for representing and propagating uncertainty in automated reasoning systems, we believe that a simple, transparent, and reasonably general formalism for eliciting joint belief functions must be developed. In particular, belief functions should admit application in problems more general than the nested partition structures studied by Shafer and Logan (1987). In this paper, we use a simple example to demonstrate that different joint belief functions may be consistent with given conditional and marginal belief functions. Several different methods for constructing a joint belief function are described, and a rationale and associated problems are considered for each.

## 2. Example

Our example concerns a company's willingness to pay expenses for a trip to the AAAI Uncertainty Workshop this summer. Either the company will pay Expenses ($E = 1$) or it will not ($E = 0$). Among other things, the company's willingness to pay will depend upon whether a paper is presented at the workshop. Let A denote Acceptance or rejection of a paper; the paper is either accepted ($A = 1$) or rejected ($A = 0$). Graphically, the relationship between these two propositions is represented by the directed link $A \to E$. These propositions actually form part of a larger system: e.g., acceptance or rejection depends on getting the paper in before the Deadline; payment of expenses impacts attendance at the Workshop ($D \to A \to E \to W$). We focus on a single link because of its simplicity, and because it is sufficient to illustrate subtleties involved in manipulating marginal and conditional belief functions in a tree-structured network. Moreover, once we understand how to propagate beliefs across a single link, further propagation is relatively straight-forward. Suppose we know how beliefs in D propagate through the link $D \to A$ to result in beliefs on A. The same mechanism can then be used to propagate the resulting beliefs in A through the next link $A \to E$ to obtain marginal beliefs in E, and so on up the chain of inference.

### 3. Three Ways of Defining Joint Beliefs

We present three possible methods for extending the conditional belief function to the joint cross-product space: conditional embedding, consonant extension, and a third method we call dissonant extension (as the resultant joint belief function is in some sense as non-consonant as possible!). For each method the marginal belief function (on A) and the conditional belief functions (on E given A) are identical.

**3.1 Conditional Embedding.** The method of conditional embedding (cf., Shafer, 1982), treats each of the conditional belief functions as an *independent* source of evidence about E. The belief function conditional on A tells us there is a .8 chance that A proves E. We represent this as a .8 belief on the set $(AE, \overline{A}E, \overline{A}\overline{E})$ (A logically implies E), and a .2 belief on the universal set $(AE, A\overline{E}, \overline{A}E, \overline{A}\overline{E})$. The belief function conditional on $\overline{A}$ tells us there is a .5 chance that $\overline{A}$ proves E. We represent this as a .5 belief on the set $(AE, A\overline{E}, \overline{A}E)$ ($\overline{A}$ logically implies E) and a .5 belief on the universal set.

Using Dempster's Rule, we combine these two belief functions with our incoming beliefs: .3 belief on $A = (AE, A\overline{E})$, .2 belief on $\overline{A} = (\overline{A}E, \overline{A}\overline{E})$, and .5 belief on the universal set. The joint belief function that results from combining these three belief functions is given in the first column of Table 1; the marginal beliefs on E are displayed in Table 2. Conditional embedding is the method suggested in Cohen, Laskey and Ulvila (1986). It is also the method that results from naive application of the Laskey and Lehner method (encoding the rules $A \rightarrow E$ and $\overline{A} \rightarrow E$ completely separately, using different belief tokens).

Table 1: Joint Belief on A x E for the 3 Propagation Mechanisms of Section 4.

| Subset | Conditional Embedding | | | Consonant Extension | | | Third Method | | |
|---|---|---|---|---|---|---|---|---|---|
| | m | Bel | Pl | m | Bel | Pl | m | Bel | Pl |
| EA | .24 | .24 | .8 | .24 | .24 | .8 | .24 | .24 | .8 |
| $\overline{E}$A | 0 | 0 | .16 | 0 | 0 | .16 | 0 | 0 | .16 |
| E$\overline{A}$ | .1 | .1 | .7 | .1 | .1 | .7 | .1 | .1 | .7 |
| $\overline{E}\overline{A}$ | 0 | 0 | .35 | 0 | 0 | .35 | 0 | 0 | .35 |
| EA, $\overline{E}$A | .06 | .3 | .8 | .06 | .3 | .8 | .06 | .3 | .8 |
| *EA, E$\overline{A}$ | .2 | .54 | 1 | .25 | .59 | 1 | .15 | .49 | 1 |
| EA, $\overline{E}\overline{A}$ | 0 | .24 | .9 | 0 | .24 | .9 | 0 | .24 | .9 |
| $\overline{E}$A, E$\overline{A}$ | 0 | .1 | .76 | 0 | .1 | .76 | 0 | .1 | .76 |
| $\overline{E}$A, $\overline{E}\overline{A}$ | 0 | 0 | .46 | 0 | 0 | .41 | 0 | 0 | .51 |
| E$\overline{A}$, $\overline{E}\overline{A}$ | .1 | .2 | .7 | .1 | .2 | .7 | .1 | .2 | .7 |
| *EA, $\overline{E}$A, E$\overline{A}$ | .05 | .65 | 1 | 0 | .65 | 1 | .1 | .65 | 1 |
| EA, $\overline{E}$A, $\overline{E}\overline{A}$ | 0 | .3 | .9 | 0 | .3 | .9 | 0 | .3 | .9 |
| *EA, E$\overline{A}$, $\overline{E}\overline{A}$ | .2 | .84 | 1 | .15 | .84 | 1 | .25 | .84 | 1 |
| $\overline{E}$A, E$\overline{A}$, $\overline{E}\overline{A}$ | 0 | .2 | .76 | 0 | .2 | .76 | 0 | .2 | .76 |
| *$\Omega_{ExA}$ | .05 | 1 | 1 | .1 | 1 | 1 | 0 | 1 | 1 |

* differences

25

Table 2: Marginal Beliefs on E for the 3 Different Propagation
Methods of Section 4.

| Subset | Conditional Embedding | | | Consonant Extension | | | Third Method | | |
|---|---|---|---|---|---|---|---|---|---|
| | m | Bel | Pl | m | Bel | Pl | m | Bel | Pl |
| E | .54 | .54 | 1 | .59 | .59 | 1 | .49 | .49 | 1 |
| $\bar{E}$ | 0 | 0 | .46 | 0 | 0 | .41 | 0 | 0 | .51 |
| $\Omega_E$ | .46 | 1 | 1 | .41 | 1 | 1 | .51 | 1 | 1 |

**3.2 Consonant Extension.** A *consonant* belief function is one in which the evidence all points in a single direction. In Shafer's terminology, the focal elements of a consonant belief function must form a nested chain of subsets. In our example, both conditional belief functions are consonant (because the focal elements {E} and {E,$\bar{E}$} form a nested sequence), but the marginal belief function over A is not (because there is no way to form a nested sequence from {A}, {$\bar{A}$}, and {A,$\bar{A}$}). If, as in our example, we are given consonant conditional belief functions on E given A and E given $\bar{A}$, there is a unique way to form a consonant joint belief function with vacuous marginals on A. (The marginal on A should be vacuous, as the conditional belief functions are meant to represent beliefs about the *link* from A to E, rather than beliefs about A itself.)

This consonant extension can be viewed as a way of representing nonindependency of the conditional belief functions. That is, in the consonant extension, we assign beliefs:

.5 focused on {AE,$\bar{A}$E}   (A proves E *and* $\bar{A}$ proves E);
.3 focused on {AE,$\bar{A}$E,$\bar{AE}$}   (A proves E but $\bar{A}$ is inconclusive);
.2 focused on {AE,A$\bar{E}$,$\bar{A}$E,$\bar{AE}$}   (both are inconclusive).

Thus, in conditional embedding we assume that A proves E *independently* of whether $\bar{A}$ proves E. In the consonant extension, we assume maximal nonindependency--the evidential links tend to be valid together. (To encode conditional embedding, Laskey and Lehner would define two rules A — E (.5) and $\bar{A}$ — E(.5) that share a belief token, and another rule A — E (.3) using another belief token) Tables 1 and 2 depict the joint and marginal beliefs obtained from combining this consonant extension with the incoming beliefs on A. Note that the marginal beliefs on E differ for the two methods. In particular, the range of permissible belief on E is smaller for the consonant extension.

**3.3 Dissonant Extension.** At one extreme of a spectrum is the nonindependency assumed by the consonant extension, where the two evidential links from A and $\bar{A}$ tend to be valid together. Conditional embedding represents a middle point, where their validity is independent. At the other extreme is the assumption that the link from A to E tends to be valid when the link from $\bar{A}$ to E is invalid, and vice versa. Making this assumption yields the following joint belief function representing the link A — E:

26

```
.3 focused on (AE,ĀE)         (A proves E and Ā proves E);
.5 focused on (AE,ĀE,AĒ)      (A proves E but Ā is inconclusive);
.2 focused on (AE,AĒ,ĀE)      (Ā proves E but A is inconclusive).
```

Note that this joint belief function has *no* belief focused on the entire cross product space (although belief *is* inconclusive with respect to E for all but the first focal element). Figures 1 and 2 depict the joint and marginal beliefs after this belief function has been combined with the incoming beliefs over A. Note that the belief range widens progressively as we move from the consonant extension to conditional embedding to our third method.

### 4  So Which Method Do I Use?

In this simple example, we believe most people's intuition would point to the results produced by the consonant extension method. Consider the following argument: "If I have .8 belief in E when A is true and .5 belief when Ā is true, then I should have at least .5 belief in E no matter which is true. The difference between .8 and .5 represents the belief that A justifies *over and above* the belief when Ā is true." Indeed, the consonant extension propagates a *vacuous* belief function over A to a belief function focusing .5 belief on (E) and .5 belief on (E,Ē).

We also note that the consonant extension method gives the same results as the natural interval probability model for this example. The reader should beware, however, of interpreting belief functions as interval probabilities--cf., Black, (1987); Laskey, (1987).

Conditional embedding, on the other hand, propagates vacuous beliefs on A to only .8×.5 = .40 belief focused on (E) and .60 belief focused on (E,Ē). This "leaking out" of belief to the universal set is a consequence of the independence assumption underlying conditional embedding. The argument for using conditional embedding would go something like this: "I can prove E *no matter what* the value of A only if *both* evidential links are valid, which has probability .8×.5 = .40."

If, as we do, you find this argument less convincing than the first, it means that you think the kind of nonindependence assumed by the consonant extension model is appropriate for this problem. Unfortunately, the consonant extension method does not generalize to the case when the conditional belief functions are not consonant. (It was no accident that the only non-consonant belief function in our example was the marginal over A). Nevertheless, this method may represent a simple and compelling way of constructing joint belief functions when the input conditional belief functions are consonant. (It should be mentioned that, like probabilities, consonant belief functions may be assessed by specifying only a single number for each hypothesis. Assessing a general belief function requires specifying a number for each member of the power set of the hypothesis space.)

We grant that the restriction to consonant belief functions may not prove a problem in many applications. However, it may be the case that a more general conditional belief function is required. The chief virtue of the method of conditional embedding is that it can be applied regardless of the structure of the conditional belief functions. If the independence assumption seems

27

untenable, the alternative is assessing a belief function over the entire cross product space, and sacrificing the simplicity of assessment based on rules.

5. <u>Caveat Modelor</u>

Extending Shafer-Dempster theory to propagating beliefs in rule-based systems requires a way to represent and elicit beliefs about the *relationship* between antecedent and consequent. Specifying belief functions on qualitative rules results in conditional belief functions, which must then be transformed into joint belief functions. We have demonstrated three different ways of extending conditional and marginal belief functions to a joint belief function. Each of these methods arises from different assumptions about the joint relationship between antecedent and consequent, and produces different results when propagating beliefs. It is our view that the method of consonant extension is the most satisfying in the example of this paper, but it is limited to consonant conditional belief functions.

The differences among methods in our simple example may seem slight, but we feel that they point to fundamental issues that deserve further study. We suspect that many readers share our uneasiness about making the kinds of independence judgments this problem asks of us. We are not sure we really understand what it means for the evidential links $A \rightarrow E$ and $\bar{A} \rightarrow E$ to be "independent" or "valid together." We believe that for belief functions to be used properly in rule-based systems, knowledge engineers and experts need to have good "canonical stories" (Shafer, 1982) that apply not just to single hypotheses but to the *links* between hypotheses.

In any case, all three analyses presented herein suggest that the company is likely to pay for the authors' trip to the conference. Readers are encouraged to substitute their own numbers to reach their own conclusions.

**References**

Adams, J.B. Probabilistic reasoning and certainty factors. In B.G. Buchanan and E.H. Shortliffe (Eds.), *Rule-based expert systems: The MYCIN experiments of the Stanford Heuristic Programming Project*, Reading, MA: Addison-Wesley Publishing Co., 1984, 263-272.

Black, P.K. Is Shafer general Bayes? *Proceedings of the Third Workshop on Uncertainty in Artificial Intelligence*, Seattle, WA, 1987, pp. 2-9.

Buchanan, B.G., and Shortliffe, E.H. *Rule-based expert systems: The MYCIN experiments of the Stanford Heuristic Programming Project*. Reading, PA: Addison-Wesley, 1984.

Cohen, M.S., Laskey, K.B., and Ulvila, J.W. *Report on methods of estimating uncertainty: Application of alternate inference theories to ABM site localization problems* (Technical Report 87-8). Falls Church, VA: Decision Science Consortium, July 1986.

Dempster, A.P. Upper and lower probabilities induced by a multivalued mapping. *Annals of Mathematical Statistics*, 1967, *38*, 325-339.

Dempster, A.P. A Generalization of Bayesian Inference (with discussion). *Journal of the Royal Statistical Society,*, Series B, 1968, *30*, 205-247.

Laskey, K.B. Belief in belief functions: An examination of Shafer's canonical examples. *Proceedings of the Third Workshop on Uncertainty in Artificial Intelligence*, Seattle, WA, 1987, pp. 39-46.

Laskey, K.B., and Lehner, P.E. Assumptions, beliefs and probabilities. *Artificial Intelligence*, in press.

Laskey, K.B., and Lehner, P. Belief maintenance: An integrated approach to uncertainty management. Presented at *American Association for Artificial Intelligence Conference*, 1988.

Lauritzen, S.L. and Spiegelhalter, D.J. Local computations with probabilities on graphical structures and their application to expert systems. *Journal of the Royal Statistical Society*, Series B, (to appear).

Pearl, J. Fusion, propagation and structuring in belief networks. *Artificial Intelligence*, 1986, *29*(3), 241-288.

Shafer, G. Belief functions and parametric models. *Journal of the Royal Statistical Society*, Series B, 1982, 44(3), 322-352.

Shafer, G. and Logan, R. Implementing Dempster's Rule for hierarchical evidence. *Artificial Intelligence*, November 1987, *33*(3), 271-298.

Shafer, G., Shenoy, P.P., and Mellouli, K. Propagating belief functions in qualitative Markov trees. *Working Paper No. 186*, Lawrence, KS: University of Kansas, School of Business, 1986.